\documentclass[runningheads]{llncs}
\usepackage{graphicx}
\usepackage{pifont}
\usepackage{color}
\usepackage[table]{xcolor}
\definecolor{LightCyan}{rgb}{0.88,1,1}
\usepackage[percent]{overpic}

\def\eg{\emph{e.g}.}

\def\etal{\emph{et al}.}

\newcommand{\Index}[2]{\index{#1, #2}#1~#2}

\begin{document}

\title{Geometry meets semantics for semi-supervised monocular depth estimation} 
\titlerunning{Geometry meets semantics for semi-supervised monocular depth estimation} 


\author{\Index{Pierluigi}{Zama Ramirez} \and
Matteo Poggi \and Fabio Tosi \and \\ 
Stefano Mattoccia \and \Index{Luigi}{Di Stefano}}
%

\authorrunning{P. Zama Ramirez et al.} 


\institute{University of Bologna, \\
Viale del Risorgimento 2, Bologna, Italy \\
\email{$\lbrace$pierluigi.zama,m.poggi,fabio.tosi5$\rbrace$@unibo.it}
}

\maketitle

\begin{abstract}
Depth estimation from a single image represents a very exciting challenge in computer vision. While other image-based depth sensing techniques leverage on the geometry between different viewpoints  (\eg, stereo or structure from motion), the lack of these cues within a single image renders ill-posed the monocular depth estimation task. For inference, state-of-the-art encoder-decoder architectures for monocular depth estimation rely on effective feature representations learned at training time. For unsupervised training of these models, geometry has been effectively exploited by suitable images warping losses computed from views acquired by a stereo rig or a moving camera. 
In this paper, we make a further step forward showing that learning semantic information from images enables to improve effectively monocular depth estimation as well. In particular, by leveraging on semantically labeled images together with unsupervised signals gained by geometry through an image warping loss, we propose a deep learning approach aimed at joint semantic segmentation and depth estimation. Our overall learning framework is semi-supervised, as we deploy groundtruth data only in the semantic domain. At training time, our network learns a common feature representation for both tasks and a novel cross-task loss function is proposed. The experimental findings show how, jointly tackling depth prediction and semantic segmentation, allows to improve depth estimation accuracy. In particular, on the KITTI dataset our network outperforms state-of-the-art methods for monocular depth estimation.
\end{abstract}

\begin{figure}
\setlength{\tabcolsep}{1pt}
\centering
\begin{tabular}{cc}
\begin{overpic}[width=0.49\textwidth]{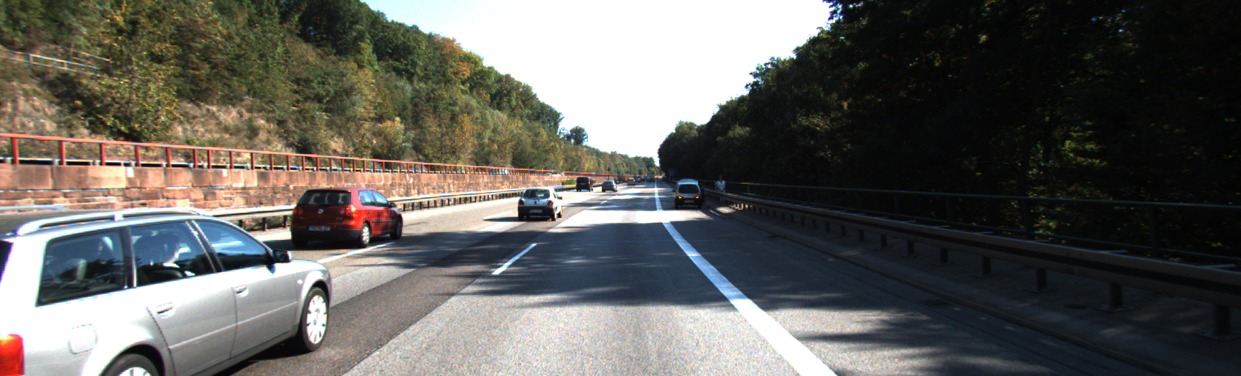}
\put (2,25) {$\displaystyle\textcolor{white}{\textbf{(a)}}$}
\end{overpic} &
\begin{overpic}[width=0.49\textwidth]{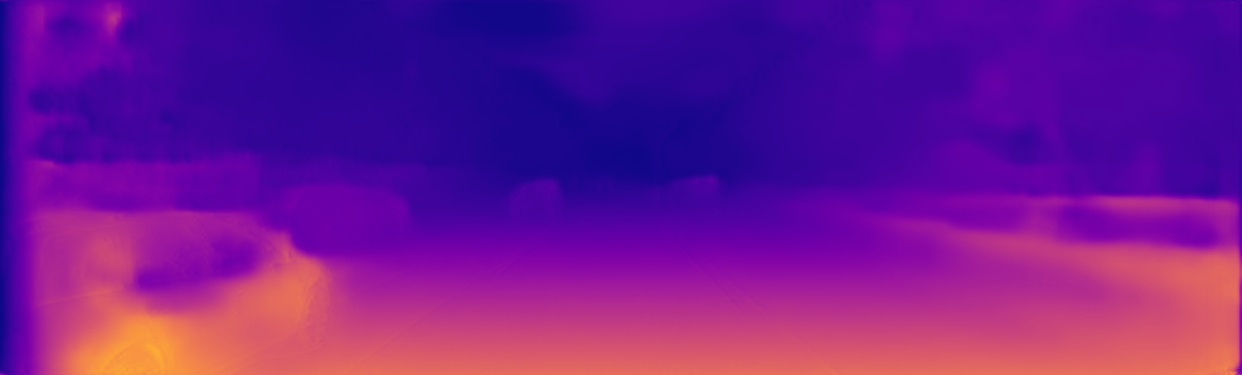}
\put (2,25) {$\displaystyle\textcolor{white}{\textbf{(b)}}$}
\end{overpic} \\
\begin{overpic}[width=0.49\textwidth]{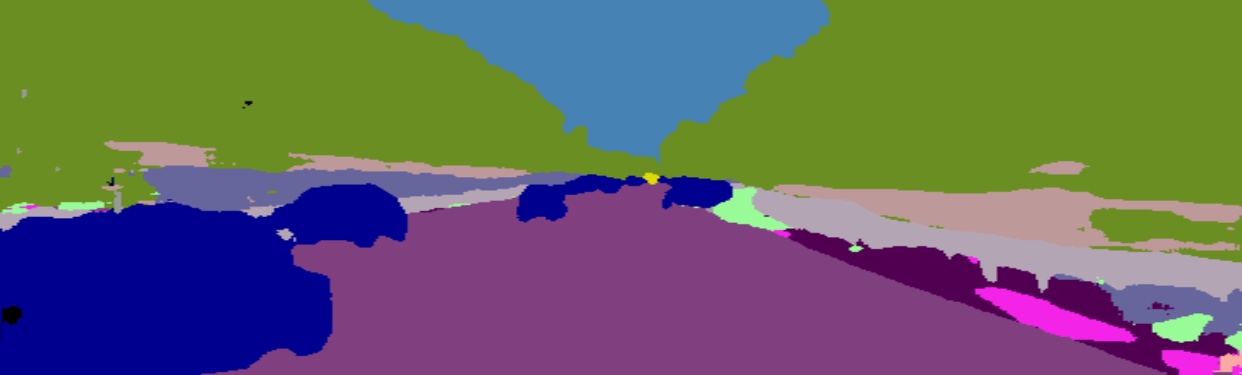}
\put (2,25) {$\displaystyle\textcolor{white}{\textbf{(c)}}$}
\end{overpic} &
\begin{overpic}[width=0.49\textwidth]{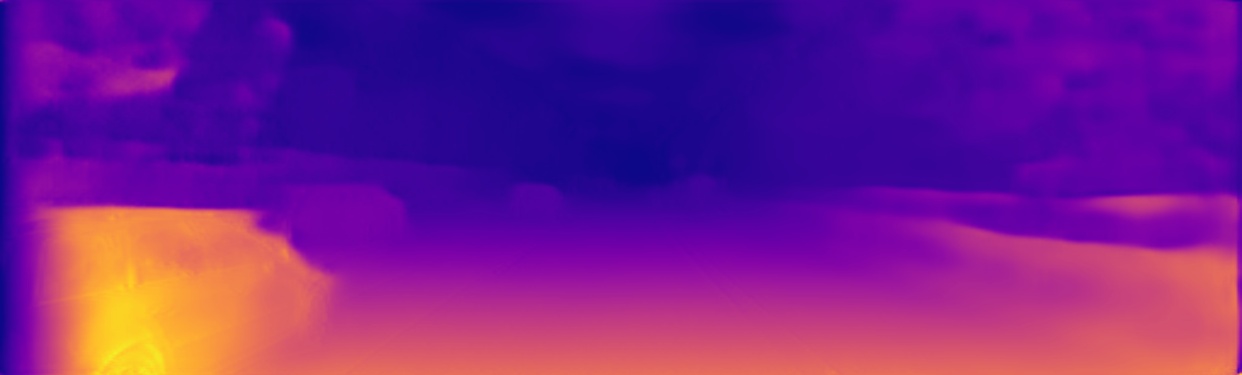}
\put (2,25) {$\displaystyle\textcolor{white}{\textbf{(d)}}$}
\end{overpic} \\
\\
\end{tabular}
\caption{Joint depth from mono and semantic segmentation. (a) Input image, (b) depth map by state-of-the-art method \cite{godard2017unsupervised}, (c) semantic and (d) depth maps obtained by our network.}
\label{fig:abstract}
\end{figure}

\section{Introduction}

Depth sensing has always played an important role in computer vision because of the increased reliability brought in by  availability of 3D data in several key tasks. In this context, dense depth estimation from images compares favorably to active sensors, such as Time-of-Flight cameras or Lidars, due to the latter either featuring short acquisition ranges or being cumbersome and much more expensive. Although traditional image-based approaches rely on multiple acquisitions from different viewpoints, like in binocular or multi-view stereo, depth estimation from a single image is receiving ever-increasing attention due to its unparalleled potential for seamless, cheap and widespread deployment.  Recently proposed supervised learning frameworks based on Convolutional Neural Networks (CNNs) have achieved excellent results on this task, though they require massive amounts of training images labeled with per pixel groundtruth depth measurements. Obtaining these labels is particularly challenging and costly as it relies on expensive active sensors, such as high-end Lidars, which typically provide sparse and noisy measurements requiring further automatic or manual processing \cite{KITTI_2012,KITTI_2015}.
To address these issues, multiple acquisitions by a stereo rig \cite{godard2017unsupervised} or a single moving camera \cite{zhou2017unsupervised} may be used to obtain supervision signals by warping different views according to the estimated depth and measuring the associated image re-projection error.

As for the depth-from-mono task, geometry cues are required at training time only. For inference, the depth estimation network is mainly driven by the learned global image context. Evidence of this can be gathered by running a monocular depth estimator, trained in either supervised or unsupervised manner, on imagery dealing with slightly different environments and observing how it may succeed in yielding reasonable results. These considerations suggest that the feature representation learned to predict depth from a single image is quite tightly linked to the \emph{semantic} content of the scene, thus it leads us to conjecture that guiding the network through explicit knowledge about scene semantics may improve effectiveness in the depth-from-mono task. 
Moreover, the very recent work by Zamir, Amir R., et al. \cite{zamir2018taskonomy}, supports the argument that learning features from multiple tasks is beneficial to performance as there exist relevant dependencies between visual tasks. Although \cite{zamir2018taskonomy} is based on fully-supervised learning, we believe that the correlation between semantic segmentation and depth estimation can be exploited also within a semi-supervised learning framework, \emph{i.e.} casting one of the two tasks in unsupervised manner.

Thus, in this paper, we propose to train a CNN architecture to perform both semantic segmentation and depth estimation from a single image. By optimizing our model jointly on the two  tasks, we enable  it to learn a more effective feature representation which yields improved depth estimation accuracy. We rely on unsupervised image re-projection loss  \cite{godard2017unsupervised}  to pursue depth prediction  whilst we let the network learn semantic information from the observed scene by supervision signals from pixel-level groundtruth semantic maps. Thus, with respect to recent work \cite{godard2017unsupervised}, our proposal requires semantically annotated imagery, thereby departing from a totally unsupervised towards a semi-supervised  learning paradigm (\emph{i.e.}  unsupervised for depth and supervised for semantics). Yet, though manual annotation of per-pixel semantic labels is tedious, it is much less prohibitive than collecting groundtruth depths. Besides, while the former task may be performed off-line after acquisition, as recently proposed for some images of the KITTI dataset \cite{Alhaija2017BMVC}, one may very unlikely obtain depth labels out of already collected frames. 

To the best of our knowledge, this paper is the first to propose integration of unsupervised monocular depth estimation with supervised semantic segmentation. By applying this novel paradigm, we improve a state-of-the-art encoder-decoder depth estimation architecture \cite{godard2017unsupervised} according to two main contributions:

\begin{itemize}

\item we propose to introduce an additional decoder stream based on the same features as those deployed for depth estimation and trained for semantic segmentation; thereby, the overall architecture is trained to optimize both tasks jointly. 

\item we propose a novel loss term, the \emph{cross-domain discontinuity} loss $\mathcal{L}_{cdd}$, aimed at enforcing spatial proximity between depth discontinuities and semantic contours.

\end{itemize}
Experimental results on the KITTI dataset prove that  tackling the two tasks jointly does improve monocular depth estimation.  For example, Fig. \ref{fig:abstract} suggests how recognizing objects like cars (c) can significantly ameliorate depth estimation (d) with respect to a depth-from-mono approach lacking any awareness about scene semantics  (b). It is also worth highlighting that, unlike all previous unsupervised frameworks in this field, our proposal  delivers not only the depth map  (Fig.\ref{fig:abstract} (d) ) but also the semantic segmentation of the input image  (Fig.\ref{fig:abstract}, (c)) by an end-to-end training process.


\section{Related work}
\label{sec:rw}

We review here the literature dealing with unsupervised monocular depth estimation and semantic segmentation, both relevant to our work.

\textbf{Unsupervised Monocular Depth.} 
Single view depth estimation \cite{liu2016learning,eigen2014depth,wang2015designing,cao2017estimating} gained much more popularity in the last years thanks to the increasing availability of benchmarks \cite{saxena2009make3d,Uhrig2017THREEDV}. Moreover, casting depth estimation as an image reconstruction task represents a very attractive way to overcome the need for expansive, groundtruth labels by using a large amount of unsupervised imagery.
The work by Garg \etal \cite{garg2016unsupervised} represents the first, pivotal step in this direction, proposing a network for monocular depth estimation by deploying, at training time, view reconstruction loss together with actual stereo pairs as supervision. Then, Godard \etal \cite{godard2017unsupervised} introduced bilinear warping \cite{jaderberg2015spatial} alongside with more robust reconstruction losses, thereby achieving state-of-the-art performance for monocular depth estimation. This approach was extended to embedded systems \cite{pydnet18}, using a virtual trinocular setup at training time \cite{3net18} or a GAN framework \cite{Aleotti_monogan_2018},  Kuznietsov \etal \cite{Kuznietsov_2017_CVPR} trained a network in a semi-supervised manner, by merging the unsupervised image reconstruction error together with the contribution from sparse depth groundtruth labels. While the above mentioned  techniques require rectified stereo pairs at training time, Zhou \etal \cite{zhou2017unsupervised} proposed to train a network to infer depth from video sequences. This network computes a reconstruction loss between subsequent frames and, at the same time, predicts the relative poses between adjacent frames. Therefore, this method enables a fully-monocular setup whereby stereo pairs are no longer required for training. However, this strategy comes to a price in performance \cite{zhou2017unsupervised}, delivering less accurate depth estimations compared to \cite{godard2017unsupervised}. 
More recent works aimed at improving the single camera supervision approach because of its easiness of use, introducing 3D point-cloud alignment \cite{mahjourian2018unsupervised}, differentiable visual odometry \cite{wang2018unsupervised}, joint optical flow estimation \cite{yin2018geonet}, or combining both stereo and video sequences supervision \cite{wang2018unsupervised}. Nevertheless, none of them actually outperforms the synergy of stereo supervision and network model deployed by Godard et al. \cite{godard2017unsupervised}.
For this reason, in this paper we follow the guidelines of \cite{godard2017unsupervised}, currently the undisputed state-of-the-art for unsupervised monocular depth estimation.

\textbf{Semantic Segmentation.}
While most early proposals relied on hand-crafted features and classifiers like Random Forests \cite{shotton2008semantic} or Support Vector Machines \cite{fulkerson2009class}, nowadays pixel-level semantic segmentation approaches mainly exploit fully convolutional neural networks \cite{long2015fully}.  Compared to previous methods, the key advantage of the present-day strategy concerns the  ability to \emph{automatically} learn a  better feature representation. Architectures for semantic segmentation focus on exploitation of contextual information and can be divided into five main groups. In the first, we find multi-scale prediction models \cite{eigen2015predicting,chen2016attention,chen2018deeplab,liang2015semantic}, whereby  the same architecture takes inputs at different scales so to extract features at different contextual levels. The second group consists of encoder-decoder architectures. The encoder is in charge of extracting low-resolution features from high-resolution inputs while the  decoder should be able to recover fine object details from the feature representation  so as to yield a high-resolution  output map \cite{long2015fully,badrinarayanan2017segnet,ronneberger2015u,lin2016refinenet}. The third group accounts for models which encode long range context information exploiting Conditional Random Fields either as a post processing module \cite{chen2018deeplab} or as an integral part of the network \cite{zheng2015conditional}. The fourth group includes  models relying on spatial pyramid pooling to extract context information at different levels \cite{zhao2017pyramid,chen2018deeplab,chen2018deeplab}. Finally, the fifth group deals with models deploying atrous-convolutions rather than the standard convolution operator to extract higher resolution features while keeping a large receptive field to capture long-range information \cite{dai2017deformable,wang2017understanding}.
Our learning framework deploys an encoder-decoder architecture to fit with the monocular depth model in \cite{godard2017unsupervised}. Thus, our semantic segmentation network may be thought of as belonging to the second group.

There exist also several works that, akin our paper, pursue joint estimation of  depth and semantics from a single image. Ladicky et al. \cite{ladicky2014pulling} combine depth regression with semantic classification deploying a bag-of-visual-words model and a boosting algorithm. Mousavian et al. \cite{mousavian2016joint} deploy a multi-scale CNN to estimate depth and used within a CRF to obtain semantic segmentation.  Wang  et al. \cite{wang2015towards} use local and global CNNs to extract pixels and regions potential which are fed  to a CRF. More recent works, such as \cite{kendall2017multi}, demonstrate that jointly performing multiple task with adequate weighting of each task can be exploited to achieve better results.  However, all these methods require groundtruth labels for both depth and semantics and are trained through multiple stages, whereas we propose to boost self-supervised depth estimation with easier to obtain semantic supervision only.

\section{Proposed method}
\label{sec:pm}
In this section, we present our proposal for joint semantic segmentation and depth estimation from a single image. We first explain  the main intuitions behind our work, then we describe the network architecture and the loss functions deployed in our deep learning framework.


Estimating the distance of objects from a camera through a single acquisition is an ill-posed problem. While other techniques can effectively measure depth based on  features extracted from different view points (\eg, binocular stereo allows for triangulating depth from point matches between two synchronized frames), monocular systems cannot rely on geometry constraints to infer distance unambiguously. Despite this lack of information, modern deep learning monocular frameworks achieved astounding results by learning effective feature representations from the observed environment. Common to latest work in this field \cite{laina2016deeper,eigen2014depth,godard2017unsupervised,zhou2017unsupervised} is the design of deep encoder-decoder architectures, with a first contractive portion progressively decimating image dimensions to reduce the computational load and increase the receptive field, followed by an expanding portion which restores the original input resolution. In particular, the encoding layers learn a high level feature representation crucial to infer depth. Although it is hard to tell what kind of information the network is actually learning at training time, we argue semantic to play an important role. Recent works like \cite{godard2017unsupervised,zhou2017unsupervised}  somehow support this intuition. Indeed, although the authors trained and evaluated their depth estimators on the KITTI dataset \cite{KITTI_2015},   a preliminary training on CityScapes \cite{cordts2016cityscapes} turned out beneficial to achieve the best accuracy with both frameworks, despite the very different camera setup between the two datasets. Common to the datasets is, in fact, the kind of sensed environment and, thus, the overall semantics of the scenes under perception. This observation represents the main rationale underpinning our proposal. By explicitly training the network to learn the semantic context of the sensed environment we shall expect to enrich the feature representation resulting from the encoding module and thus obtain a more accurate depth estimation.  This may be realized by a deep model in which a single encoder is shared between two decoders in charge of providing, respectively, a depth map and a semantic segmentation map. Accordingly,  minimization of the errors with respect to pixel-level semantic labels provides gradients  that flow back into the encoder at training time, thereby learning a shared feature representation aware of both depth prediction  as well as scene semantics.  According to our claim, this should turn out conducive to better depth prediction. 

\begin{figure}[t]
\centering\includegraphics[width=0.90\textwidth]{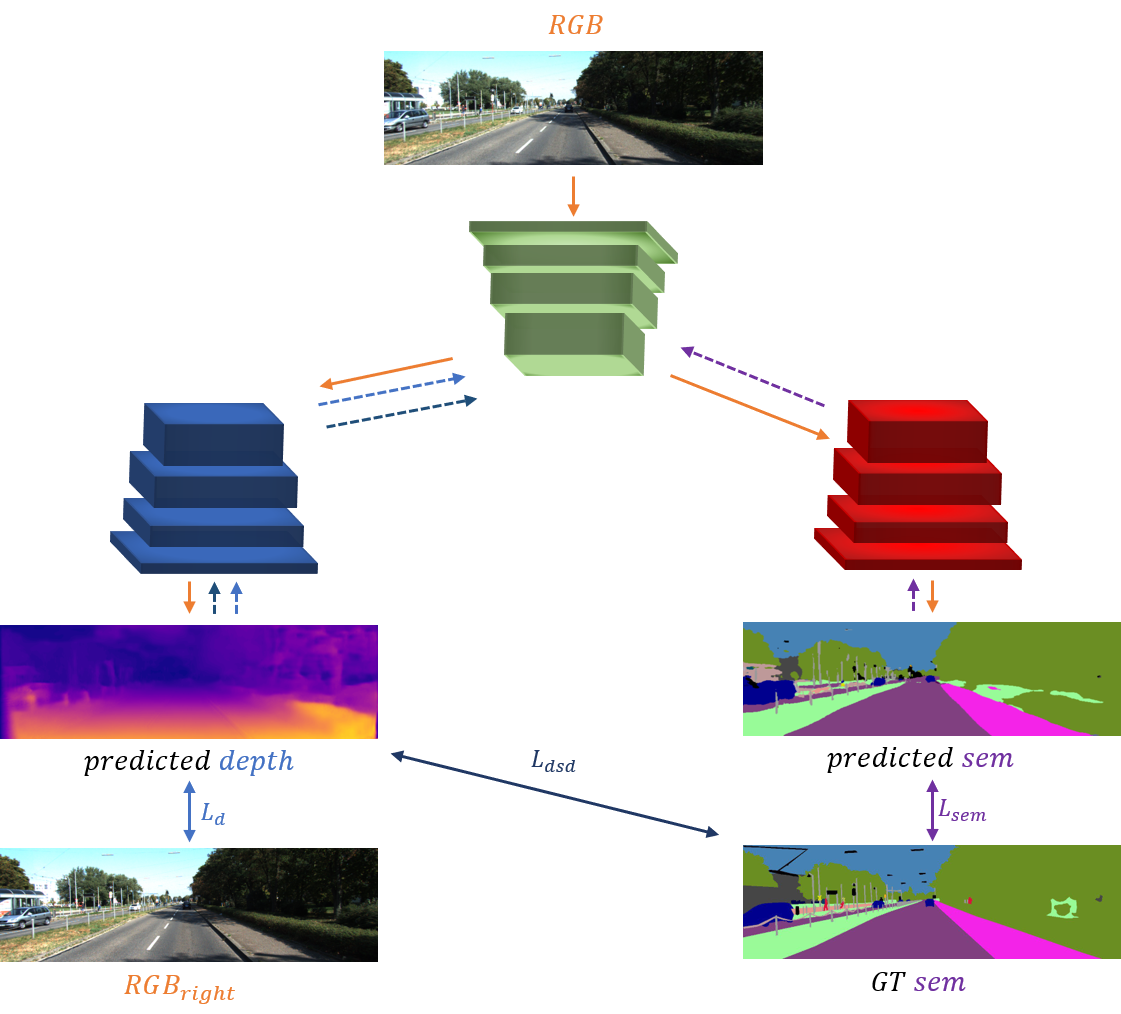}
\caption{Schematic representations of the proposed network architecture and semi-supervised learning framework. A single encoder (green) is shared between a depth (blue) and a semantic (red) decoder. The depth decoder is  optimized to minimize $\mathcal{L}_d$ and $\mathcal{L}_{cdd}$,  the semantic decoder to minimize $\mathcal{L}_s$. }
\label{fig:conciseArchitecture}
\end{figure}

Inspired by successful attempts to predict depth from a single image, we design a suitable encoder-decoder architecture for joint depth estimation and semantic segmentation. The encoder  is in charge of learning a rich feature representation by increasing the receptive field of the network while reducing the input dimension and computational overhead. Popular encoders for this task are VGG \cite{Simonyan2014VeryDC} and ResNet50 \cite{he2016deep} . The decoder  restores the original input resolution by means of up-sampling operators followed by $3 \times 3$ convolutions linked  by means of skip connections with the encoder at the corresponding resolution. As illustrated  in Fig. \ref{fig:conciseArchitecture},  to  infer both depth and semantics we keep relying on a single encoder (green) and  replicate the decoder  to realize a second estimator. The two decoders (blue, red) do not share weights and are trained to minimize different losses, which deal with the depth prediction (blue) and semantic segmentation (red) tasks. While the two decoders are updated by different gradients flows, the shared encoder (green)  is updated according to both flows, thereby learning a representation optimized  jointly for the two  tasks.
We validate our approach by extending the architecture proposed by  Godard \etal \cite{godard2017unsupervised} for monocular depth estimation: the encoder produces two inverse depth (i.e., disparity) maps by processing the left image of a stereo pair. Then, the right image is used to obtain supervision signals by warping the left image according to the estimated disparities, as explained in the following section. 

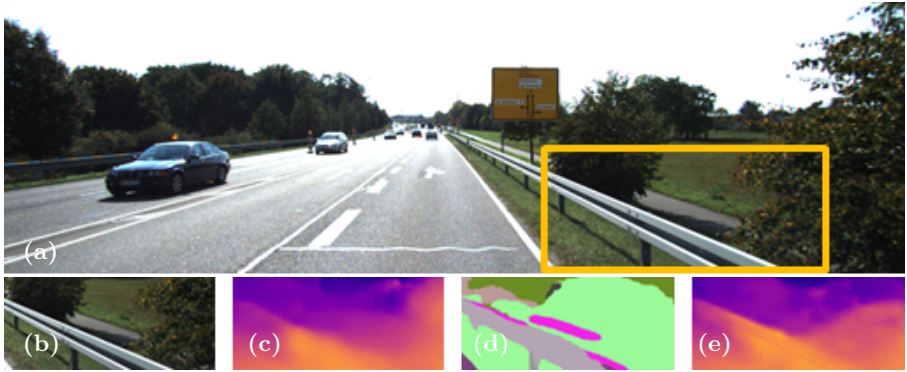
\begin{figure}[t]
\setlength{\tabcolsep}{1pt}
\centering
\begin{overpic}[width=0.98\textwidth]{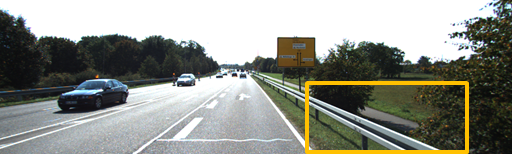}
\put (2,2) {$\displaystyle\textcolor{white}{\textbf{(a)}}$}
\end{overpic}
\begin{overpic}[width=0.98\textwidth]{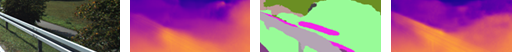}
\put (2,2) {$\displaystyle\textcolor{white}{\textbf{(b)}}$}
\put (27,2) {$\displaystyle\textcolor{white}{\textbf{(c)}}$}
\put (52,2) {$\displaystyle\textcolor{white}{\textbf{(d)}}$}
\put (77,2) {$\displaystyle\textcolor{white}{\textbf{(e)}}$}
\end{overpic}
\caption{Example of improved depth estimation enabled by semantic knowledge. (a) input image, (b) region extracted from the scene, (c) depth map predicted by \cite{godard2017unsupervised}, depth (d) and semantic (e) maps predicted by our framework. We can clearly notice how the the structure of the guard rail is better preserved by our method (e) compared to \cite{godard2017unsupervised} in (c).}
\label{fig:detail}
\end{figure}

Figure \ref{fig:detail} shows how the shared representation used to jointly tackle both tasks enables to reconstruct better shapes when estimating depth (e) thanks to the semantic context (d) learned by the network compared to standalone learning of depth (c) as in \cite{godard2017unsupervised}. 

\subsection{Loss functions}

To train the proposed architecture, we rely on the following multi-task loss function 

\begin{equation}
\mathcal{L}_{tot} = \alpha_{d} \mathcal{L}_{d} + \alpha_{s} \mathcal{L}_{s} + \alpha_{cdd} \mathcal{L}_{cdd}
\label{eq:totloss}
\end{equation}

which consists in the weighted sum of three terms, namely the  \emph{depth} ($\mathcal{L}_d$), \emph{semantic} ($\mathcal{L}_s$) and \emph{cross-domain discontinuity} ($\mathcal{L}_{cdd}$) terms. As shown in in Fig. \ref{fig:conciseArchitecture}, each term back-propagates gradients through a different decoder:  in particular,  $\mathcal{L}_d$ and $\mathcal{L}_{cdd}$  through the depth (blue) decoder whilst $\mathcal{L}_s$  through the semantic (red) decoder. All gradients then converge so to flow back into the shared (green) encoder.

\subsubsection{Depth term}

The depth term, $\mathcal{L}_d$, in our multi-task loss is computed according to the unsupervised training paradigm proposed by Godard \etal \cite{godard2017unsupervised}: 

\begin{equation}
\mathcal{L}_d = \beta_{ap}(\mathcal{L}^l_{ap} + \mathcal{L}^r_{ap}) + \beta_{ds}(\mathcal{L}^l_{ds}+\mathcal{L}^r_{ds}) + \beta_{lr}(\mathcal{L}^l_{lr}+\mathcal{L}^r_{lr})
\label{eq:loss}
\end{equation}

where the loss consists in the weighted sum of three terms, namely  the \emph{appearance}, \emph{disparity smoothness} and \emph{left-right consistency} terms. The first term  measures the image re-projection error by means of the SSIM \cite{wang2004image} and L1 difference between the original and warped images, $I$ and $\tilde{I}$:

\begin{equation}
\mathcal{L}^l_{ap} = \frac{1}{N} \sum_{i,j} \gamma \frac{1 - SSIM(I^l_{i,j},\tilde{I}^l_{i,j})}{2} + (1-\gamma)||(I^l_{i,j}-\tilde{I}^l_{i,j})||
\end{equation}

The smoothness term penalizes large disparity differences between neighboring pixels along the $x$ and $y$ directions unless these occur in presence of strong intensity gradients in the reference image $I$

\begin{equation}
\mathcal{L}^l_{ds} = \frac{1}{N} \sum_{i,j} |\delta_x d^l_{i,j}|e^{-||\delta_x I^l_{i,j}||} + |\delta_y d^l_{i,j}|e^{-||\delta_y I^l_{ij}||}
\end{equation}

Finally, the left-right consistency enforces coherence between the predicted disparity maps, $d^l$ and $d^r$, for left and right images: 

\begin{equation}
\mathcal{L}^l_{lr} = \frac{1}{N} \sum_{i,j} |d^l_{i,j} - d^r_{i,j + d^l_{i,j}}|
\end{equation}

As proposed in \cite{godard2017unsupervised}, in our learning framework $\mathcal{L}_d$ is computed at four different scales.

\subsubsection{Semantic term}

The semantic term $\mathcal{L}_s$ within our total loss is given by the standard cross-entropy between the predicted and groundtruth pixel-wise semantic labels: 

\begin{equation}
\mathcal{L}_s = \mathcal{C}(p_t,\overline{p}_t) = H({p}_t,\overline{p}_t) + KL(p_t,\overline{p}_t)
\end{equation}

where $H$ denotes the entropy and $KL$ the  $KL-$divergence.  The semantic term,  $\mathcal{L}_s$, is computed at full resolution only.

\subsubsection{Cross-domain discontinuity term}

We also introduce a novel cross-task loss term aimed at enforcing an explicit link between the two learning tasks by leveraging on the groundtruth pixel-wise semantic labels to improve depth prediction. We found that the most effective manner to realize this consists in deploying the observation that depth discontinuities are likely to co-occur with semantic boundaries. Accordingly, we have designed the following \emph{cross-domain discontinuity}, $\mathcal{L}_{cdd}$, term: 

\begin{equation}
\mathcal{L}_{cdd} = \frac{1}{N} \sum_{i,j} sgn(|\delta_x sem^{l}_{i,j}|)e^{-||\frac{\delta_x d^l_{i,j}}{d^l_{i,j}}||}+  sgn(|\delta_y sem^{l}_{i,j}|)e^{-||\frac{\delta_y d^l_{i,j}}{d^l_{i,j}}||}
\end{equation}

where $\textit{sem}$ denotes the ground truth semantic map and $\textit{d}$  the predicted disparity map. Differently from the smoothness term $\mathcal{L}^l_{ds}$ in the disparity domain, the novel $\mathcal{L}_{cdd}$ term detects discontinuities between semantic labels encoded by the sign of the absolute value of the gradients in the semantic map. The idea behind this loss is that there should be a gradient peak between adjacent pixels belonging to different classes. Nevertheless, we do not care about its magnitude since the numeric labels do not have any mathematical meaning. 

\section{Experimental results}

In this section, we compare the performance of our semi-supervised joint depth estimation and semantic segmentation paradigm with respect to the proposal by Godard \etal \cite{godard2017unsupervised}, which represents nowadays the undisputed state-of-the-art for unsupervised monocular depth estimation. As discussed in Sec. \ref{sec:pm}, our method as well as the baseline used in our experiments, \emph{i.e.} \cite{godard2017unsupervised}, require rectified stereo pairs at training time. Suitable datasets for this purpose are thus CityScapes \cite{cordts2016cityscapes} and KITTI \cite{KITTI_RAW}, which provide a large number of training samples, \emph{i.e.} about 23k and 29k rectified stereo pairs respectively. However, our method requires also pixel-wise groundtruth semantic labels at training time, which limits the actual amount of training samples available for our experiments.  In particular, CityScapes includes about 3k finely annotated images, while the KITTI 2015 benchmark recently made available pixel-wise semantic groundtruths for about 200 images \cite{Alhaija2017BMVC}.
Therefore, to carry out a fair evaluation of the actual contribution provided by semantic information in the depth-from-mono task to the baseline fully unsupervised approach, we trained both methods based on the reduced datasets featuring stereo pairs alongside with semantically annotated left frames.

\subsection{Implementation details}

Our proposal has been implemented in Tensorflow\footnote{Source code and trained models are available at \url{https://github.com/CVLAB-Unibo/Semantic-Mono-Depth}.}, starting from the source code made available by the authors of \cite{godard2017unsupervised}.
We adhere to the original training protocol by Godard \etal, scheduling 50 epochs on the CityScapes dataset and 50 further on the KITTI 2015 images. For quantitative evaluation, we split the KITTI 2015 dataset into train and test sets, providing more details in the next section.
We train on 256$\times$512 images using a batch dimension of 2, we set the previously introduced hyper-parameters as follows: $\alpha_d=1$, $\alpha_s=0.1$, $\alpha_{cdd}=0.1$, $\beta_{ap}=1$, $\beta_{lr}=1$, $\beta_{ds}=\frac{1}{r}$ (being $r$ the down-sampling factor at that resolution) and $\gamma=0.85$. 
Models are trained using Adam optimizer \cite{kingma2014adam}, with $\beta_1=0.9$, $\beta_1=0.999$ and $\epsilon=10^{-8}$. The initial learning rate is set to $10^{-4}$, halved after 30 and 40 epochs.
We perform data augmentation on input RGB images, in particular random gamma, brightness and color shifts sampled within the ranges [0.8,1.2] for gamma, [0.5,2.0] for brightness, and
[0.8,1.2] for each color channel separately. 
Moreover we flip images horizontally with a probability of 50\%. If the flip occurs, the right image in the stereo pair becomes the new reference image and we do not provide supervision signals from semantics (as  right semantic maps are not available in the datasets).
We implemented our network with both VGG and ResNet50 encoders, as in \cite{godard2017unsupervised}. The semantic decoder adds about 20.5M parameters, resulting in nearly 50 and 79 million parameters for the two models (31 and 59, respectively, for \cite{godard2017unsupervised}).

\subsection{Monocular depth estimation: evaluation on KITTI 2015}

We quantitatively assess the effectiveness of our proposal on the KITTI 2015 training dataset for stereo \cite{KITTI_2015}. It provides 200 synchronized pairs of images together with groundtruth disparity and semantic maps \cite{Alhaija2017BMVC}. 
As already mentioned, to carry out a fair comparison between our approach and \cite{godard2017unsupervised}, we can use only these samples and thus the numerical results reported in our paper cannot be compared directly with those in \cite{godard2017unsupervised}. Then, we randomly split the 200 pairs from KITTI into 160 training samples and 40 samples used only for evaluation\footnote{The testing samples, belonging to the KITTI 2015 dataset, are: 000001, 000003, 000004, 000019, 000032, 000033, 000035, 000038, 000039, 000042, 000048, 000064, 000067, 000072, 000087, 000089, 000093, 000095, 000105, 000106, 000111, 000116, 000119, 000123, 000125, 000127, 000128, 000129, 000134, 000138, 000150, 000160, 000161, 000167, 000174, 000175, 000178, 000184, 000185 and 000193.}. We measure the accuracy of the predicted depth maps after training for 50 epochs on CityScapes and then fine-tuning for 50 more epochs on the samples selected from KITTI.

Table \ref{tab:ablation} reports quantitative results using VGG or ResNet50 as backbone encoder. Each model, one per row in the table, is trained with four different strategies:
\begin{itemize}
    \item 
    $\mathcal{L}_d$ uses only the depth term as loss (\emph{i.e.}, equivalently to the baseline approach by Godard \etal \cite{godard2017unsupervised}).
    \item
    $\mathcal{L}_d$+$\mathcal{L}_s$ adds the semantic term to the depth term.
    \item
    $\mathcal{L}_d$+$\mathcal{L}_s$+$\mathcal{L}_{cdd}$ minimizes our proposed total loss function (Equation \ref{eq:totloss}).
    \item
    $\mathcal{L}_d$+$\mathcal{L}_{cdd}$ minimizes only the losses dealing with the depth decoder.
\end{itemize}
The table provides results yielded by the four considered networks according to standard performance evaluation metrics \cite{godard2017unsupervised} computed between estimated depth $d$ and groundtruth $D$.

This ablation highlights how introducing the second decoder trained to infer semantic segmentation maps, significantly improves depth prediction according to all performance metrics for both type of encoder. Moreover, adding the cross-domain discontinuity term, $\mathcal{L}_{cdd}$, leads in most cases to further improvements. On the other hand, minimizing $\mathcal{L}_d$ and $\mathcal{L}_{cdd}$ alone leads to inferior performance compared to the baseline method. 
We obtain the best configuration according to all metrics using ResNet50 when both $\mathcal{L}_s$ and $\mathcal{L}_{cdd}$ are minimized alongside with the depth term $\mathcal{L}_d$. 



\begin{table*}[t]
\center
\begin{tabular}{|c|cc|cccc|ccc|}
\cline{6-9}
\multicolumn{5}{c}{} & \multicolumn{2}{|c|}{\cellcolor{blue!25}Lower is better}
 & \multicolumn{2}{c|}{\cellcolor{LightCyan}Higher is better} & \multicolumn{1}{c}{} \\
\hline
 & Encoder & pp & \cellcolor{blue!25}Abs Rel & \cellcolor{blue!25}Sq Rel & \cellcolor{blue!25}RMSE & \cellcolor{blue!25}RMSE log & \cellcolor{LightCyan}$\delta_1$ & \cellcolor{LightCyan}$\delta_2$ & \cellcolor{LightCyan}$\delta_3$\\
\hline
$\mathcal{L}_d$ \cite{godard2017unsupervised} & VGG &  & 0.160 & 2.707 & 7.220 & 0.239 & 0.837 & 0.928 & 0.966 \\
$\mathcal{L}_d$+$\mathcal{L}_s$ & VGG &  & 0.155 & 2.511 & 6.968 & \textbf{0.234} & 0.841 & \textbf{0.931} & \textbf{0.968} \\
$\mathcal{L}_d$+$\mathcal{L}_s$+$\mathcal{L}_{cdd}$ & VGG & & \textbf{0.154} & \textbf{2.453} & \textbf{6.949} & 0.235 & \textbf{0.844} & \textbf{0.931} & 0.967\\
$\mathcal{L}_d$+$\mathcal{L}_{cdd}$ & VGG &  & 0.161 & 2.758 & 7.128 & 0.240 & 0.841 & 0.928 & 0.964 \\
\hline
$\mathcal{L}_d$ \cite{godard2017unsupervised} & VGG & \ding{51} & 0.149 &  2.203 &  6.582 & 0.223 & 0.844 & 0.936 & 0.972 \\
$\mathcal{L}_d$+$\mathcal{L}_s$ & VGG & \ding{51} & 0.147 & 2.229 & 6.583 & 0.223 & 0.847 & \textbf{0.938} & \textbf{0.972} \\
$\mathcal{L}_d$+$\mathcal{L}_s$+$\mathcal{L}_{cdd}$ & VGG & \ding{51} & \textbf{0.145} & \textbf{2.040} & \textbf{6.362} & \textbf{0.221} & \textbf{0.849} & \textbf{0.938} & 0.971\\
$\mathcal{L}_d$+$\mathcal{L}_{cdd}$ & VGG & \ding{51} & 0.150 & 2.278 & 6.539 & 0.225 & 0.843 & 0.934 & 0.970 \\
\hline
\hline
$\mathcal{L}_d$ \cite{godard2017unsupervised} & ResNet &  & 0.159 & 2.411 & 6.822 & 0.239 & 0.830 & 0.930 & 0.967 \\
$\mathcal{L}_d$+$\mathcal{L}_s$ & ResNet &  &  0.152 & 2.385 & 6.775 & 0.231 & 0.843 & 0.934 & 0.970 \\
$\mathcal{L}_d$+$\mathcal{L}_s$+$\mathcal{L}_{cdd}$ & ResNet &  & \textbf{0.143} & \textbf{2.161} & \textbf{6.526} & \textbf{0.222} & \textbf{0.850} & \textbf{0.939} & \textbf{0.972}\\
$\mathcal{L}_d$+$\mathcal{L}_{cdd}$ & ResNet &  & 0.155 & 2.282 & 6.658 & 0.232 & 0.840 & 0.932 & 0.968 \\
\hline
$\mathcal{L}_d$ \cite{godard2017unsupervised} & ResNet & \ding{51} & 0.148 & 2.104 & 6.439 & 0.224 & 0.839 & 0.936 & 0.972 \\
$\mathcal{L}_d$+$\mathcal{L}_s$ & ResNet & \ding{51} & 0.144 & 2.050 & 6.351 & 0.220 & 0.849 & 0.938 & 0.972 \\
$\mathcal{L}_d$+$\mathcal{L}_s$+$\mathcal{L}_{cdd}$ & ResNet & \ding{51} & \textbf{0.136} & \textbf{1.872} & \textbf{6.127} & \textbf{0.210} & \textbf{0.854} & \textbf{0.945} & \textbf{0.976}\\
$\mathcal{L}_d$+$\mathcal{L}_{cdd}$ & ResNet & \ding{51} & 0.144 & 1.973 & 6.199 & 0.217 & 0.849 & 0.940 & 0.975 \\
\hline
\end{tabular}
\caption{Ablation experiments on KITTI 2015 evaluation split, using different configurations of losses, encoders and post-processing (pp). Best setup highlighted in bold for each configuration.}
\label{tab:ablation}
\end{table*}

Moreover, we also evaluated the output obtained by all models after performing the post-processing step proposed by \cite{godard2017unsupervised}, that consists in forwarding both the input image $I$ and its horizontally flipped counterpart $\hat{I}$. This produces two depth maps $d_I$ and $d_{\hat{I}}$, the latter is flipped back obtaining $\hat{d}_{\hat{I}}$ and averaged with the former, in order to reduce artifacts near occlusions.  We can notice that the previous trend is confirmed. In particular, the full loss $\mathcal{L}_d + \mathcal{L}_s + \mathcal{L}_{cdd}$ leads to the best result on most scores. Furthermore, including the post-processing step allows the VGG model trained with our full loss to outperform the baseline ResNet50 architecture supervised by traditional depth losses only. This fact can be noticed in Table \ref{tab:ablation} comparing row 7 with row 13, observing that the former leads to better results except for $\delta_3$ metric.

\begin{table*}[t]
\center
\begin{tabular}{|c|cccc|ccc|}
\cline{4-7}
\multicolumn{3}{c}{} & \multicolumn{2}{|c|}{\cellcolor{blue!25}Lower is better}
 & \multicolumn{2}{c|}{\cellcolor{LightCyan}Higher is better} & \multicolumn{1}{c}{} \\
\hline
 & \cellcolor{blue!25}Abs Rel & \cellcolor{blue!25}Sq Rel & \cellcolor{blue!25}RMSE & \cellcolor{blue!25}RMSE log & \cellcolor{LightCyan}$\delta_1$ & \cellcolor{LightCyan}$\delta_2$ & \cellcolor{LightCyan}$\delta_3$\\
\hline
Zhou et al. \cite{zhou2017unsupervised} & 0.286 & 7.009 & 8.377 & 0.320 & 0.691 & 0.854 & 0.929 \\
Mahjourian et al. \cite{mahjourian2018unsupervised} & 0.235 & 2.857 & 7.202 & 0.302 & 0.710 & 0.866 & 0.935 \\
Yin et al. \cite{yin2018geonet} & 0.236 & 3.345 & 7.132 & 0.279 & 0.714 & 0.903 & 0.950 \\
Godard et al. \cite{godard2017unsupervised} & 0.159 & 2.411 & 6.822 & 0.239 & 0.830 & 0.930 & 0.967 \\
Ours & \textbf{0.143} & \textbf{2.161} & \textbf{6.526} & \textbf{0.222} & \textbf{0.850} & \textbf{0.939} & \textbf{0.972}\\
\hline
\end{tabular} 
\caption{Comparison with other self supervised method on KITTI 2015 evaluation split. Both \cite{godard2017unsupervised} and our method use ResNet50 encoder.}
\label{tab:comparison}
\end{table*}

To further prove the effectiveness of our proposed method we compare it with other self-supervised approach as \cite{yin2018geonet},\cite{zhou2017unsupervised},\cite{mahjourian2018unsupervised}. Thus, we have ran experiments with the source code available from \cite{yin2018geonet},\cite{zhou2017unsupervised},\cite{mahjourian2018unsupervised} using the same testing data as for \cite{godard2017unsupervised} and our method. Table \ref{tab:comparison} shows the outcome of this evaluation. We point out that we used the weights made available by the authors of \cite{yin2018geonet},\cite{zhou2017unsupervised},\cite{mahjourian2018unsupervised}, trained on a much larger amount of data (i.e., the entire Cityscapes and KITTI sequences, some of them overlapping with the testing split as well) w.r.t. the much lower supervision provided to our network. Despite this fact, monocular supervised works \cite{yin2018geonet},\cite{zhou2017unsupervised},\cite{mahjourian2018unsupervised} perform poorly compared to both [1] and our approach, confirming our semi-supervised framework to outperform them as well. We also point out that our test split relies on high-quality ground-truth labels for evaluation, available from KITTI 2015 stereo dataset, while the Eigen split used to validate \cite{yin2018geonet},\cite{zhou2017unsupervised},\cite{mahjourian2018unsupervised} provides much worse quality depth measurements, as also argued by the authors of \cite{godard2017unsupervised}.

As our final test we also compare our method with the recent multi-task learning approach by Kendall et al. \cite{kendall2017multi}. Differently from our approach, they jointly learn depth, semantic and instance segmentation in fully supervised manner. They run experiments Tiny Cityscapes, a split obtained by resizing the validation set of Cityscapes to 128 × 256 resolution. To compare our results to theirs we have taken our ResNet50 model trained on Cityscapes and validated it following the same protocol. Their depth-only model (trained supervised) achieves 0.640 inverse mean depth error, dropping to 0.522 when trained to tackle semantic and instance segmentation as well. Our ResNet50 network (trained unsupervised) starts with 1.705 error for depth-only, dropping to 1.488. Thus, the two approaches achieve 22\% and 15 \% improvement respectively. We point out that, besides relying on supervised learning for depth, \cite{kendall2017multi} exploits both semantic and instance segmentation, requiring additional manually annotated labels, while we only enforce our cross-domain discontinuity loss.


\begin{figure*}[t]
\setlength{\tabcolsep}{1pt}
\centering
\begin{tabular}{cc}
\begin{overpic}[width=0.48\textwidth]{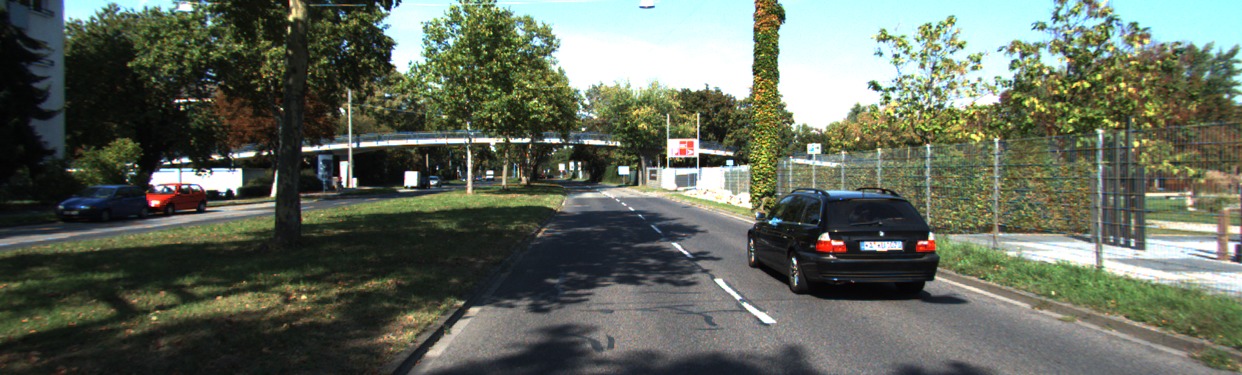}
\put (2,25) {$\displaystyle\textcolor{white}{\textbf{(a)}}$}
\end{overpic} &
\begin{overpic}[width=0.48\textwidth]{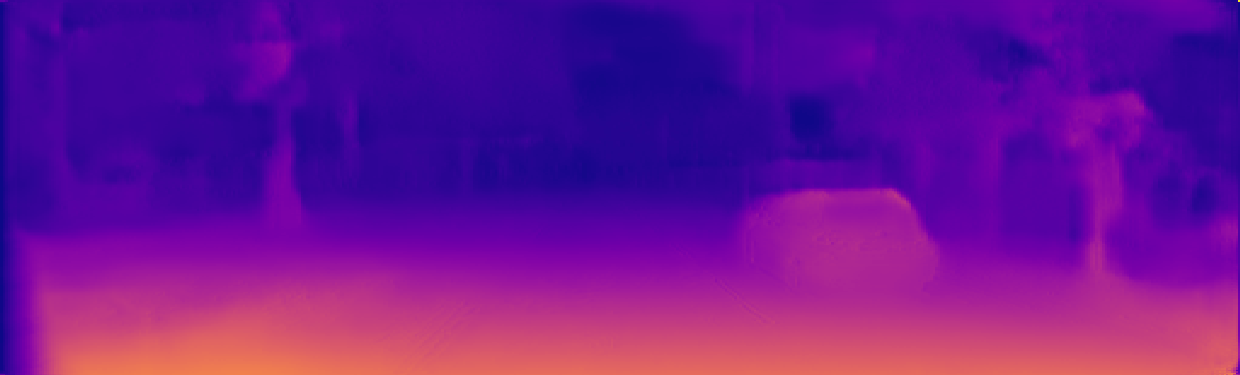}
\put (2,25) {$\displaystyle\textcolor{white}{\textbf{(b)}}$}
\end{overpic} \\
\begin{overpic}[width=0.48\textwidth]{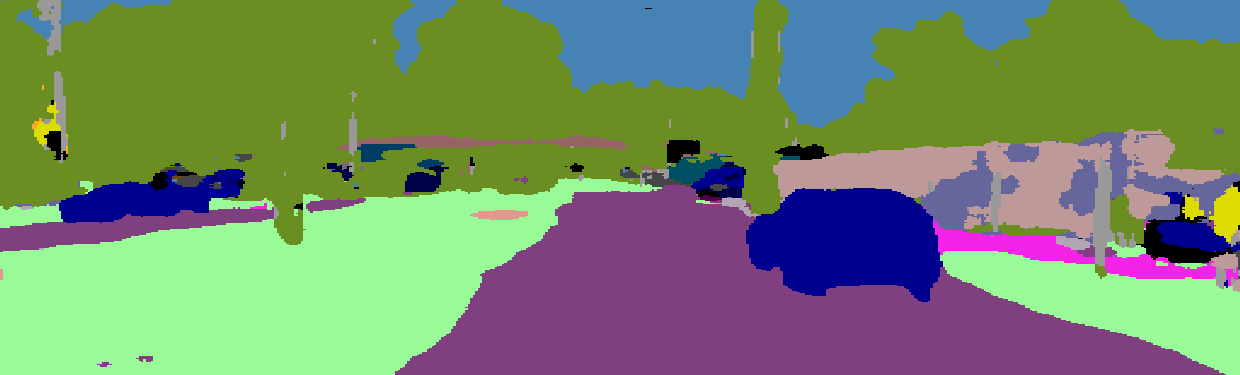}
\put (2,25) {$\displaystyle\textcolor{white}{\textbf{(c)}}$}
\end{overpic} &
\begin{overpic}[width=0.48\textwidth]{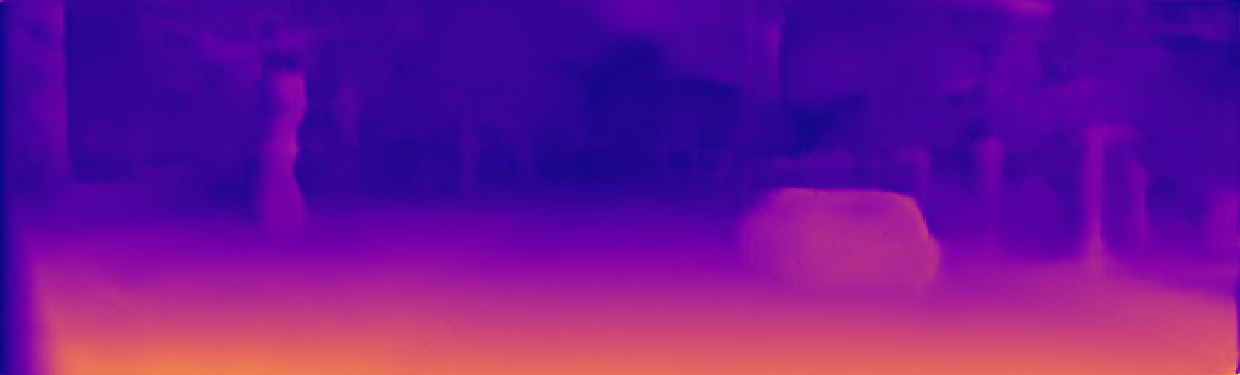}
\put (2,25) {$\displaystyle\textcolor{white}{\textbf{(d)}}$}
\end{overpic} \\
\begin{overpic}[width=0.48\textwidth]{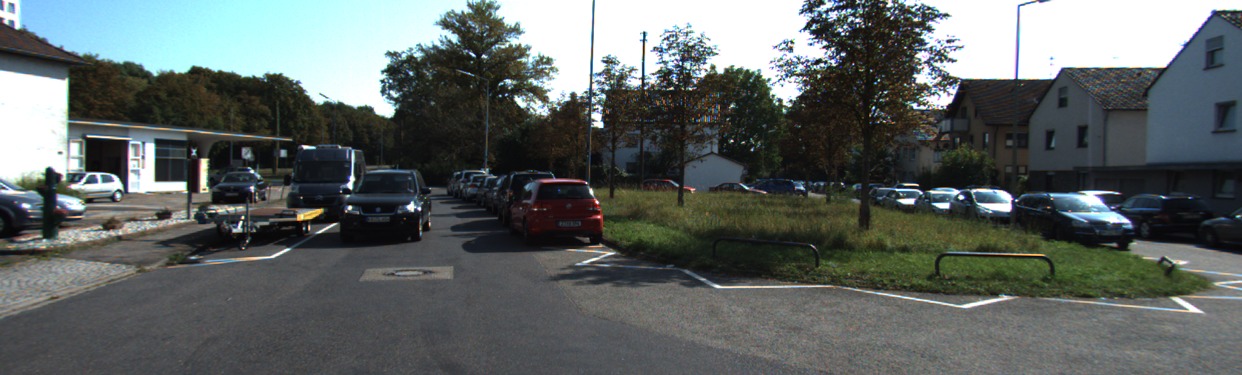}
\put (2,25) {$\displaystyle\textcolor{white}{\textbf{(a)}}$}
\end{overpic} &
\begin{overpic}[width=0.48\textwidth]{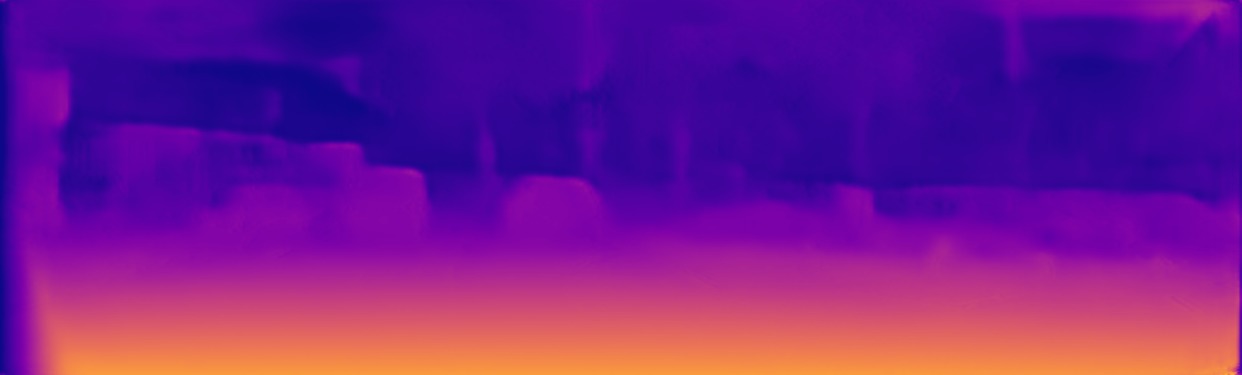}
\put (2,25) {$\displaystyle\textcolor{white}{\textbf{(b)}}$}
\end{overpic} \\
\begin{overpic}[width=0.48\textwidth]{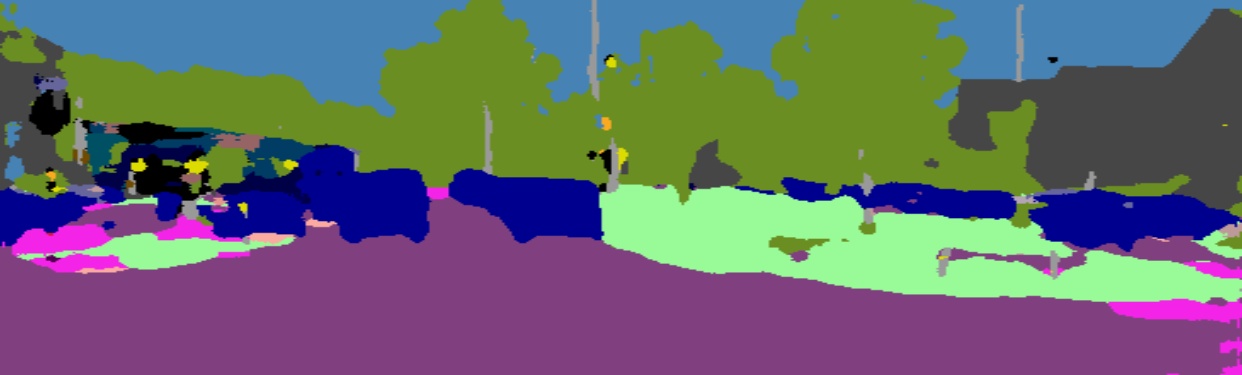}
\put (2,25) {$\displaystyle\textcolor{white}{\textbf{(c)}}$}
\end{overpic} &
\begin{overpic}[width=0.48\textwidth]{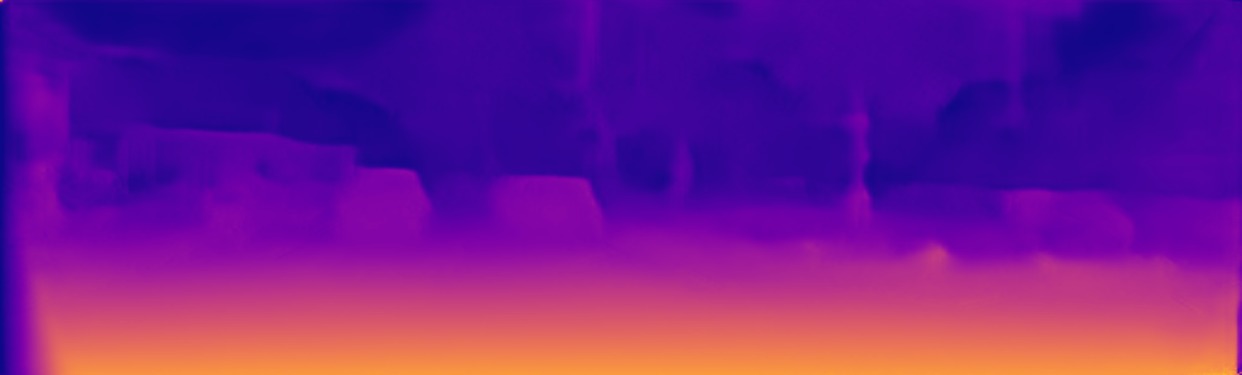}
\put (2,25) {$\displaystyle\textcolor{white}{\textbf{(d)}}$}
\end{overpic} \\

\begin{overpic}[width=0.48\textwidth]{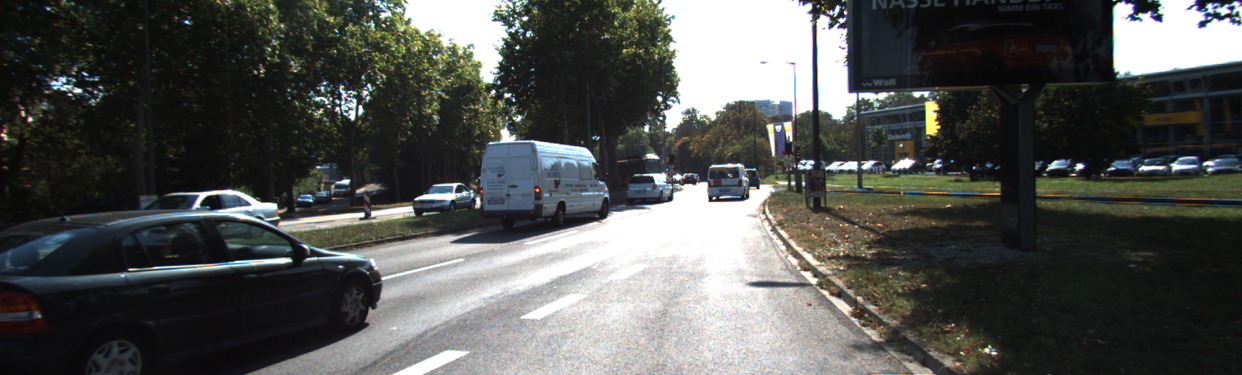}
\put (2,25) {$\displaystyle\textcolor{white}{\textbf{(a)}}$}
\end{overpic} &
\begin{overpic}[width=0.48\textwidth]{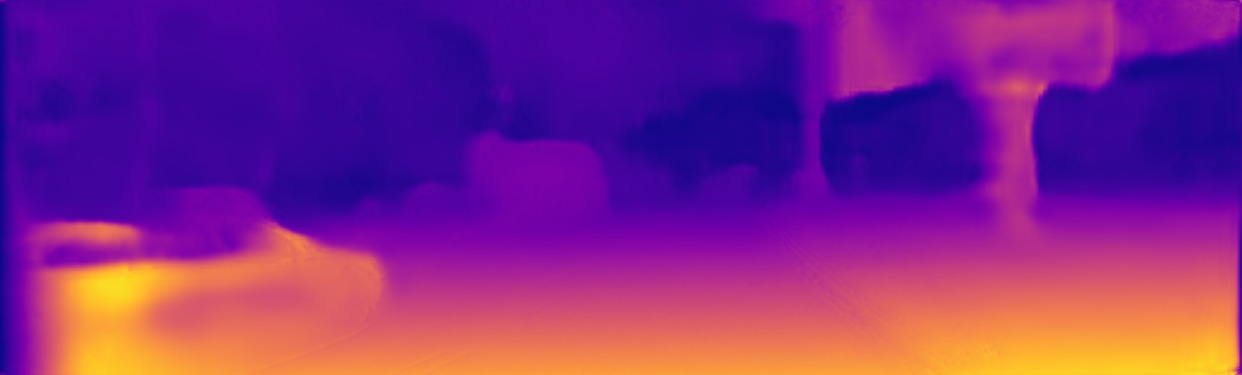}
\put (2,25) {$\displaystyle\textcolor{white}{\textbf{(b)}}$}
\end{overpic} \\
\begin{overpic}[width=0.48\textwidth]{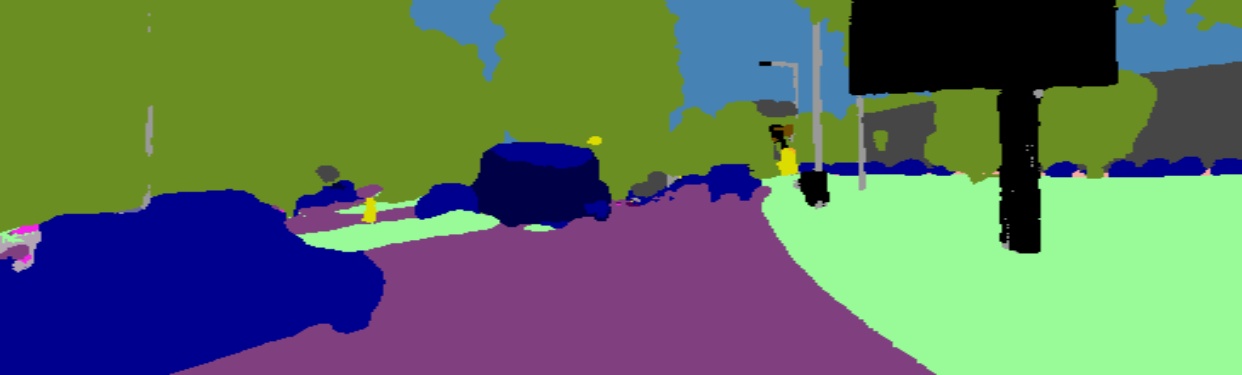}
\put (2,25) {$\displaystyle\textcolor{white}{\textbf{(c)}}$}
\end{overpic} &
\begin{overpic}[width=0.48\textwidth]{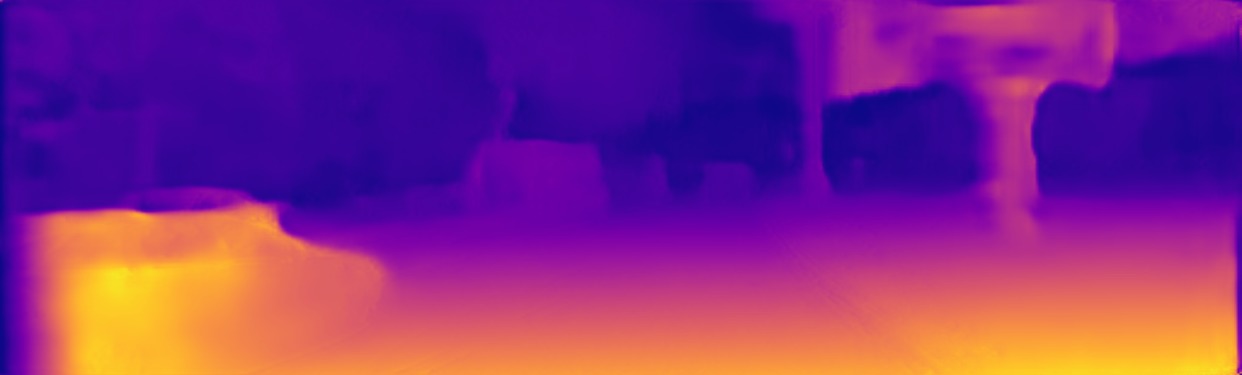}
\put (2,25) {$\displaystyle\textcolor{white}{\textbf{(d)}}$}
\end{overpic} \\
\\
\end{tabular}
\caption{Qualitative comparison between \cite{godard2017unsupervised} and our proposal on KITTI 2015 evaluation split \cite{KITTI_2015}. (a) Input image, (b) depth map by \cite{godard2017unsupervised}, (c) and (d) semantic and depth maps by our approach. Both models use Resnet50 as encoder. From top to bottom, results concerning images 000019, 000087 and 000095 belonging to our evaluation split.}
\label{fig:qualitative}
\end{figure*}

Figure \ref{fig:qualitative} depicts a qualitative comparison between the depth maps predicted by \cite{godard2017unsupervised} and our semi-supervised framework. In the figure, from top to bottom, we consider images 000019 and 000095 belonging to our evaluation split. We can observe how explicitly learning the semantics of the scene helps to correct wrong depth estimations, especially on challenging objects.
For example, we can notice how depth maps predicted by our frameworks provide better car shapes thanks to the contribution given by the semantic. This fact is particularly evident in correspondence of reflective or transparent surfaces like car windows as reported on image 000095. Moreover, the quality of thin structures like poles is improved as well, as clearly perceivable by looking at frame 000019.

\subsection{Semantic segmentation: evaluation on KITTI 2015}

Although our proposal is aimed at ameliorating depth prediction by learning richer features exploiting semantics, our network also delivers a semantic segmentation of the input image. To gather hints about the accuracy of this additional outcome of our network, we evaluated the semantic maps generated on the same KITTI evaluation split defined before. Differently from the monocular depth estimation task, results concerning semantic segmentation are quite far from the state-of-the-art. 
In particular, we obtain 88.51\% and 88.19\% per-pixel accuracy, respectively, with models based on VGG and ResNet50. We ascribe this to our architecture - inspired by \cite{godard2017unsupervised} - being optimized for unsupervised depth prediction, whereas different design choices are often found in networks pursuing semantic segmentation (i.e., atrous convolutions, SPP layers ...). We also found that training the basic encoder-decoder for semantic segmentation only yields to 86.72\% and 88.18\% per-pixel accuracy with VGG and ResNet50, respectively. Thus, while semantics helps depth prediction inasmuch as to outperform the state-of-the-art within the proposed framework, the converse requires further studies as the observed improvements are indeed quite minor. Therefore, we plan to investigate on how to design a network architecture and associated semi-supervised learning framework whereby the synergy between monocular depth prediction and semantic segmentation may be exploited in order to significantly improve accuracy in both tasks. 

\section{Conclusion and future work}

We have proposed a deep learning architecture to improve unsupervised monocular depth estimation by leveraging on semantic information. We have shown how training our architecture end-end to infer semantics and depth jointly enables us to outperform the state-of-the-art approach for unsupervised monocular depth estimation \cite{godard2017unsupervised}. Our single-encoder/dual-decoder architecture is trained in a semi-supervised manner, \emph{i.e.} using ground truth labels only for the semantic segmentation task. 
Despite obtaining groundtruth labels for semantic is tedious and requires accurate and time-consuming manual annotation, it is still more feasible than depth labeling. In fact, this latter task requires expensive active sensors to be used at acquisition time and becomes almost unfeasible offline, on already captured frames. Thus, our method represents an attractive alternative to improve self-supervised training without adding more image samples.
Future work will i) explore single camera sequences as supervision \cite{zhou2017unsupervised,yin2018geonet,wang2018unsupervised} and ii) dig into the semantic segmentation side of our framework, to reach top accuracy on this second task as well and to propose a state-of-the-art framework for joint depth and semantic estimation in a semi-supervised manner. 

\textbf{Acknowledgments} We gratefully acknowledge the support of NVIDIA Corporation
with the donation of the Titan X GPU used for this research

\bibliographystyle{splncs}
\bibliography{egbib}

\end{document}